\documentclass[conference]{IEEEtran}
\IEEEoverridecommandlockouts
\usepackage{graphicx}
\usepackage{setspace}
\usepackage{amsmath}
\usepackage{dsfont}
\usepackage{amsfonts}
\usepackage{lipsum}
\usepackage{multicol}
\usepackage{float}
\usepackage{bm}
\usepackage{color}
\usepackage[dvipsnames]{xcolor}
\usepackage{etoolbox}
\usepackage{soul}
\usepackage[linesnumbered,ruled,vlined]{algorithm2e}
\usepackage{mathtools}
\usepackage{caption}
\usepackage{amsthm}
\usepackage{adjustbox}
\usepackage{afterpage}
\usepackage{multirow}
\usepackage{mathrsfs}
\usepackage{hyperref}
\usepackage{oplotsymbl}
\usepackage{booktabs}
\usepackage{array}
\usepackage{xcolor}
\usepackage{amssymb}
\usepackage[caption = false,subrefformat=parens,labelformat=parens]{subfig}

\title{When to Stop Federated Learning: Zero-Shot Generation of Synthetic Validation Data\\with Generative AI for Early Stopping

\author{
Youngjoon Lee\textsuperscript{\rm 1}\textsuperscript{\dag}, Hyukjoon Lee\textsuperscript{\rm 2}, Jinu Gong\textsuperscript{\rm 3}, Yang Cao\textsuperscript{\rm 4}, and Joonhyuk Kang\textsuperscript{\rm 1} \\
\textsuperscript{\rm 1}School of Electrical Engineering, KAIST, South Korea\\
\textsuperscript{\rm 2}AI Group, AMD, United States\\
\textsuperscript{\rm 3}Department of Applied AI, Hansung University, South Korea\\
\textsuperscript{\rm 4}Department of Computer Science, Institute of Science Tokyo, Japan\\
Email: yjlee22@kaist.ac.kr, jkang@kaist.ac.kr
}
\thanks{This research was supported by the Institute of Information \& Communications Technology Planning \& Evaluation (IITP)-ITRC (Information Technology Research Center) grant funded by the Korea government (MSIT)  (No.RS-2025-02309685, Development of Programmable
Infrastructure Technology for Guaranteed Application Performance).\\
\textsuperscript{\dag} Work done while a visiting student at Institute of Science Tokyo.
}

}

\begin{document}

\maketitle

\begin{abstract}
Federated Learning (FL) enables collaborative model training across decentralized devices while preserving data privacy. 
However, FL methods typically run for a predefined number of global rounds, often leading to unnecessary computation when optimal performance is reached earlier.
In addition, training may continue even when the model fails to achieve meaningful performance.
To address this inefficiency, we introduce a zero-shot synthetic validation framework that leverages generative AI to monitor model performance and determine early stopping points. 
Our approach adaptively stops training near the optimal round, thereby conserving computational resources and enabling rapid hyperparameter adjustments. 
Numerical results on multi-label chest X-ray classification demonstrate that our method reduces training rounds by up to 74\% while maintaining accuracy within 1\% of the optimal.
\end{abstract}

\section{Introduction}\label{sec:intro}
Deep learning has achieved remarkable performance across numerous applications, yet it heavily relies on centralized data collection and processing \cite{li2020federated}. 
In practice, increasing concerns about data privacy and security regulations pose significant challenges to traditional centralized approaches \cite{zhao2018federated}.
Federated Learning (FL) addresses these privacy issues by training models collaboratively across decentralized devices without sharing local data directly \cite{zhou2018convergence, lee2022accelerated}. 
In FL, a central server aggregates model updates computed locally by participating devices, iteratively improving the global model through multiple global rounds \cite{li2019convergence}. 
This decentralized learning paradigm effectively maintains data privacy while leveraging distributed data sources to enhance model performance.

However, practical deployment of FL in real-world scenarios reveals significant limitations related to computational resource management \cite{kairouz2021advances}. 
Specifically, FL methods typically run for a predefined number of global rounds, regardless of whether optimal model performance is reached earlier. 
This wastes computation once peak performance is reached, undermining the cost-effectiveness of real-world FL deployments \cite{10643330}.
Additionally, performance heavily depends on key hyperparameters whose near-optimal settings may not be detectable until all predefined rounds are completed \cite{lee2025revisit,lee2025debunking}.

\begin{figure}[t]
     \centering
     \includegraphics[width=\columnwidth]{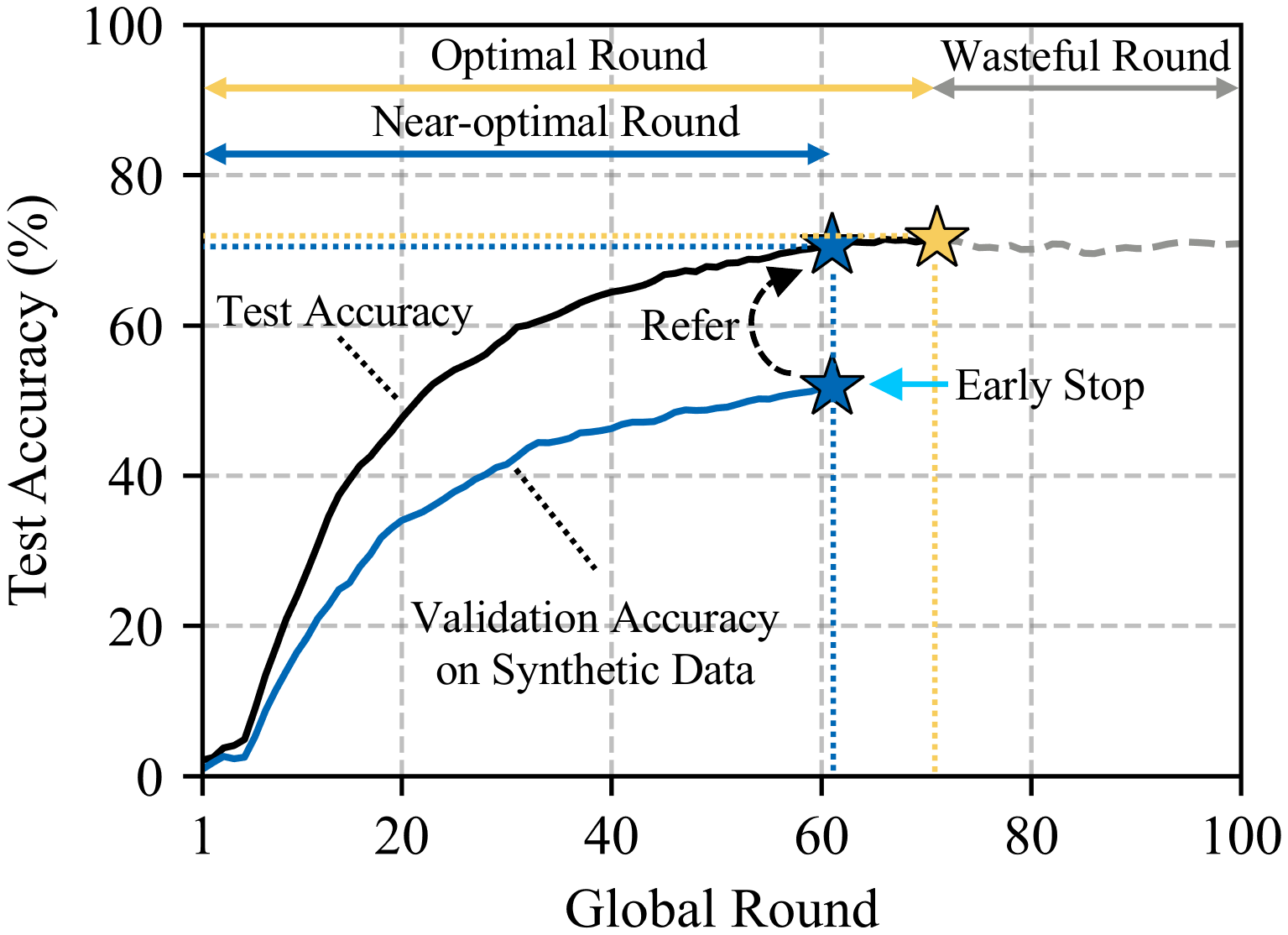}
     \vspace{-0.5cm}
     \caption{Illustration of our proposed synthetic validation-based early stopping approach in FL. The central server monitors validation accuracy on synthetic data (\textcolor[HTML]{0068B5}{blue line}) to determine the early stopping point (\textcolor[HTML]{0068B5}{$\starletfill$}) near the optimal round, achieving comparable performance to the actual optimal round (\textcolor[HTML]{F7CD5D}{$\starletfill$}) while avoiding wasteful computational rounds.}
     \label{fig:fig1}
     
\end{figure}

\begin{figure*}[t]
     \centering
     \includegraphics[width=\textwidth]{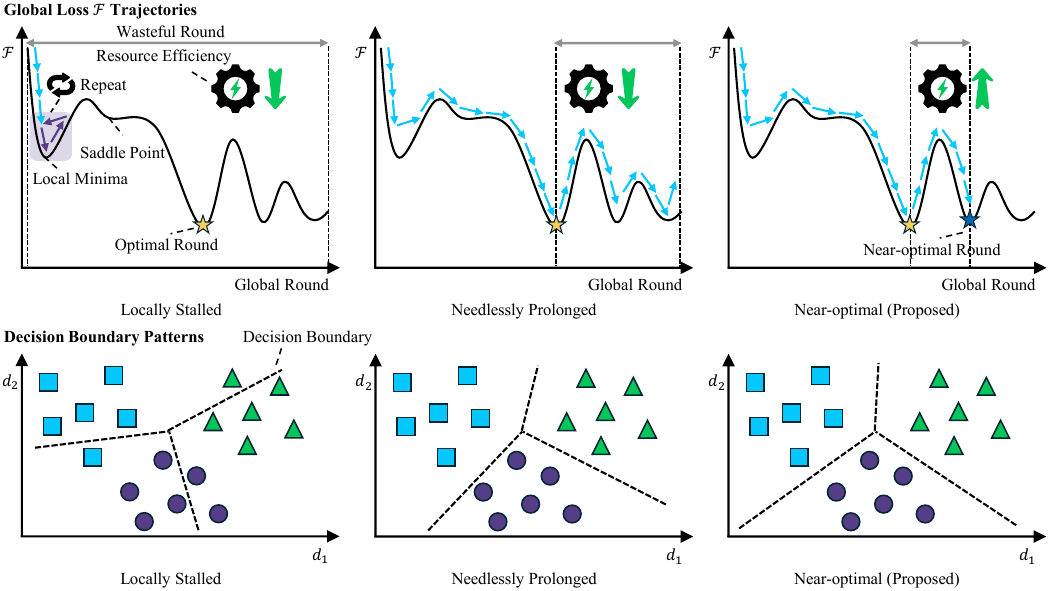}
     \caption{Illustration of global loss trajectories and corresponding decision boundary patterns across different federated training scenarios.
     Note that each arrow denotes the aggregation of each global round in FL.
     }
    \label{fig:fig3}
\end{figure*}

To address these challenges, we propose an early stopping approach that leverages generative AI to create synthetic validation datasets in a zero-shot manner, as illustrated in Fig.~\ref{fig:fig1}.
Our approach enables the central server to monitor model performance throughout federated training using synthetically generated validation sets. 
By continuously evaluating the global model on synthetic data, our method identifies near-optimal stopping points, eliminating wasteful rounds, as depicted in Fig.~\ref{fig:fig3}.
Furthermore, this approach not only conserves computational resources by avoiding unnecessary training rounds but also enables rapid evaluation and adjustment of hyperparameter settings.
Thus, our proposed method improves resource utilization efficiency and accelerates the optimization process in FL. 

The main contributions of this paper are as follows:
\begin{itemize}
\item We propose an early stopping framework that utilizes generative AI-driven synthetic datasets as a validation set within FL, enabling efficient resource usage.
\item We numerically show that our framework stops at rounds comparable to the optimal points of each state-of-the-art FL method, cutting unnecessary computation.
\item In addition, we validate that using a foundation model fine-tuned on similar domain enables faster early stopping compared to vanilla foundation models.
\end{itemize}

The remainder of this paper is organized as follows.
In Section~\ref{sec:main}, we introduce the FL setting and our synthetic validation–based early stopping framework.
Section~\ref{sec:experiment} provides results on chest X-ray classification with analyses across non-IID degrees and ablation studies.
Finally, Section~\ref{sec:conclusion} concludes with remarks.

\section{Problem and Model}\label{sec:main}
\subsection{Federated Setting}
We consider a FL setting involving $N$ participating devices and a central server following the standard FL protocol. 
Each device $i \in {1, 2, \ldots, N}$ possesses a local chest X-ray dataset $D^i_{real}$ containing medical images with corresponding multi-label diagnostic annotations. 
To reflect non-IID nature of FL, we employ a label skew setting \cite{li2022federated} that follows a Dirichlet distribution to simulate heterogeneous pathological condition distributions across devices in the global data union $D_g = \bigcup_{i=1}^N D^i_{real}$.
The objective is to collaboratively learn a global model $w_g$ for multi-label chest X-ray  while ensuring that sensitive medical data remains decentralized and private. 

Accordingly, FL aims to minimize the global empirical risk function:
\begin{equation}
\min_{w_g} [\mathcal{F}(w_g)] = \frac{1}{N} \sum_{i=1}^N \mathcal{F}_i(w_g),
\end{equation}
where $\mathcal{F}(w_g)$ represents the global objective function aggregated at the central server. Each local objective function $\mathcal{F}_i(w_i)$ for device $i$ is defined as:
\begin{equation}
\mathcal{F}_i(w_i) = \frac{1}{|D^i_{real}|} \sum{(x_j, y_j) \in D^i_{real}} \ell(w_i; x_j, y_j),
\end{equation}
where $\ell(\cdot)$ denotes the binary cross-entropy with logits loss function suitable for multi-label classification tasks, and $(x_j, y_j)$ represents an image paired with its corresponding multi-label.
Note that, while we focus on multi-label classification with binary cross-entropy loss, our proposed synthetic-validation-based early stopping framework is fundamentally generalizable.
The method can naturally extend to other machine learning tasks such as single-label classification, regression, and segmentation, provided suitable generative models exist to produce synthetic validation datasets.

\subsection{Proposed Framework}
We propose a general early stopping framework applicable to any FL method, leveraging synthetic validation data generated via generative AI.
To incorporate our proposed early stopping framework naturally into FL workflow, we outline the detailed process below within the standard FL steps.
At first, before FL training begins, the central server synthesizes a validation dataset $D_{syn}$ through generative AI employing zero-shot prompt engineering.
Each synthetic sample $(\tilde{x}_j, \tilde{y}_j)$ is generated using explicit textual prompts that describe the targeted diagnostic labels:
\begin{equation}
\text{prompt} = \text{"Frontal chest X-ray with } {\text{target label}}\text{"}.
\end{equation}
The quality of the synthetic data critically determines the accuracy and reliability of our proposed early stopping strategy. 
To ensure the quality of synthetic data, we employ advanced generative models, specifically Stable Diffusion variants (v1.4, v1.5, v2.0, and XL) and RoentGen, a generative model fine-tuned specifically for medical chest X-ray images.
Note that, the synthetic validation dataset $D_{syn} = \bigcup_{j=1}^M{(\tilde{x}_j, \tilde{y}_j)}$,
remains fixed throughout the training process. 

After generating the synthetic validation data, the central server begins FL by broadcasting the global model $w_g^{r}$ to all participating devices.
Initially, at round $r=0$, the global model is randomly initialized. 
Subsequently, the server randomly selects a subset of devices, denoted as $\mathcal{S}_r$, comprising $K \ll N$ devices from the total $N$, to participate actively in the current round. Only these selected devices engage in the subsequent local update, while the rest remain idle until the next communication cycle.

Upon receiving the global model, each selected device $k \in \mathcal{S}_r$ initializes its local model parameters $w^r_{k}$ with the received global parameters $w_g^{r}$. 
Each device then performs local training on its dataset $D_{real}^{k}$ for up to a maximum number of local rounds, updating its local model parameters via an FL-specific optimization routine:
\begin{equation}
w^r_k \leftarrow EdgeOpt(w_g^{r}, D_{real}^{k}),
\end{equation}
where the $EdgeOpt(\cdot)$ may vary based on the FL method employed (e.g., standard local SGD in FedAvg or advanced methods incorporating additional hyperparameters).

After the completion of local training, the central server collects the updated models $\{ w_k^r \}_{k=1}^K$ from the participating devices and aggregates as:
\begin{equation}
w_g^{r+1} \leftarrow ServerOpt(\{ w_k^r \}_{k=1}^K),
\end{equation}
where $ServerOpt(\cdot)$ varies by FL method.
At this point, our approach introduces synthetic validation data to evaluate the updated global model. Specifically, the central server computes the synthetic validation accuracy $ValAcc_{syn}(\cdot)$:
\begin{equation}
ValAcc_{syn}(w_g^{r+1}) = \frac{1}{|D_{syn}|} \sum_{(\tilde{x}_j, \tilde{y}_j) \in D_{syn}} \mathbf{1}[f(w_g^{r+1}; \tilde{x}_j) = \tilde{y}_j],
\end{equation}
where $\mathbf{1}[\cdot]$ is the indicator function and $f(\cdot)$ represents the model's prediction function.

Based on the computed synthetic validation accuracy, the server identifies the near-optimal early stopping round $r_{near}^{*}$ as the earliest round after the predefined minimum patience threshold $p$ at which no positive relative improvement has been observed for $p$ consecutive rounds:
\begin{equation}
r_{near}^{*} = \min \left\{ r \geq p \;\middle|\; \Delta^{\,r+1-\tau} \leq 0,\;\forall \tau \in \{1, 2, \dots, p\} \right\},
\end{equation}
where the relative improvement at round $r+1$ is defined as:
\begin{equation}
\Delta^{r+1} = \frac{ValAcc_{syn}(w_g^{r+1}) - ValAcc_{syn}(w_g^{r})}{ValAcc_{syn}(w_g^{r})}.
\end{equation}
Otherwise, the process repeats for the next round.
The overall procedure is described in Algorithm 1.

Integrating this framework into FL workflows significantly enhances computational efficiency by ensuring that training halts near optimal performance without unnecessary computational overhead. 
Notably, our approach keeps the core FL principle that sensitive real data remains exclusively on devices, thus preserving the decentralized and privacy-preserving characteristics of FL.

\begin{algorithm}[h]
\caption{Proposed Synthetic Validation–Based Early Stopping in FL}
\label{alg:earlystop}
\DontPrintSemicolon
\SetAlgoVlined
\SetKwInOut{Input}{Input}\SetKwInOut{Output}{Output}
\SetKwProg{Fn}{Function}{}{}

\Input{$N$ devices with $\{D_{\text{real}}^{k}\}$; $K$ participants; patience $p$; max rounds $R_{\max}$; generator $G$; class set $\mathcal{C}$;}
\Output{Early-stopped global model $w_g^{\,r^{*}_{\text{near}}}$}

\BlankLine
\textbf{Step 1: Synthetic validation set} \;
$D_{\text{syn}} \gets \bigcup_{c \in \mathcal{C}}\{ G(\text{``Frontal chest X-ray with }c\text{''}) \}$ \;

\BlankLine
\textbf{Step 2: Federated training with early stopping} \;
Initialize $w_g^0$, $V \gets \textsc{Evaluate}(D_{\text{syn}}, w_g^0)$, $\kappa \gets 0$ \;
\textit{/* Global model initialized randomly */} \;
\For{$r \gets 0$ \KwTo $R_{\max}-1$}{
  Sample $\mathcal{S}_r \subset \{1,\dots,N\}, |\mathcal{S}_r|=K$ \;
  Each $k \in \mathcal{S}_r$: $w_k^r \gets \text{EdgeOpt}(w_g^r, D_{\text{real}}^{k})$ \;
  $w_g^{r+1} \gets \text{ServerOpt}(\{w_k^r\}_{k\in\mathcal{S}_r})$ \;
  $V' \gets \textsc{Evaluate}(D_{\text{syn}}, w_g^{r+1})$ \;
  \eIf{$V' \le V$}{ $\kappa \gets \kappa + 1$ }{ $\kappa \gets 0$ }
  \If{$r+1 \ge p$ \textbf{ and } $\kappa = p$}{ $r^{*}_{\text{near}} \gets r+1$; \textbf{break} }
  $V \gets V'$ \;
}
\textit{/* Stop when patience $p$ is reached without improvement */} \;
\Return $w_g^{\,r^{*}_{\text{near}}}$ \;

\BlankLine
\Fn{\textsc{Evaluate}($D_{\text{syn}}$, $w$)}{%
  \Return $\tfrac{1}{|D_{\text{syn}}|}\sum_{(\tilde{x},\tilde{y})\in D_{\text{syn}}}\mathbf{1}[f(w;\tilde{x})=\tilde{y}]$ \;
}
\end{algorithm}

\begin{figure*}[t]
     \centering
     \includegraphics[width=\textwidth]{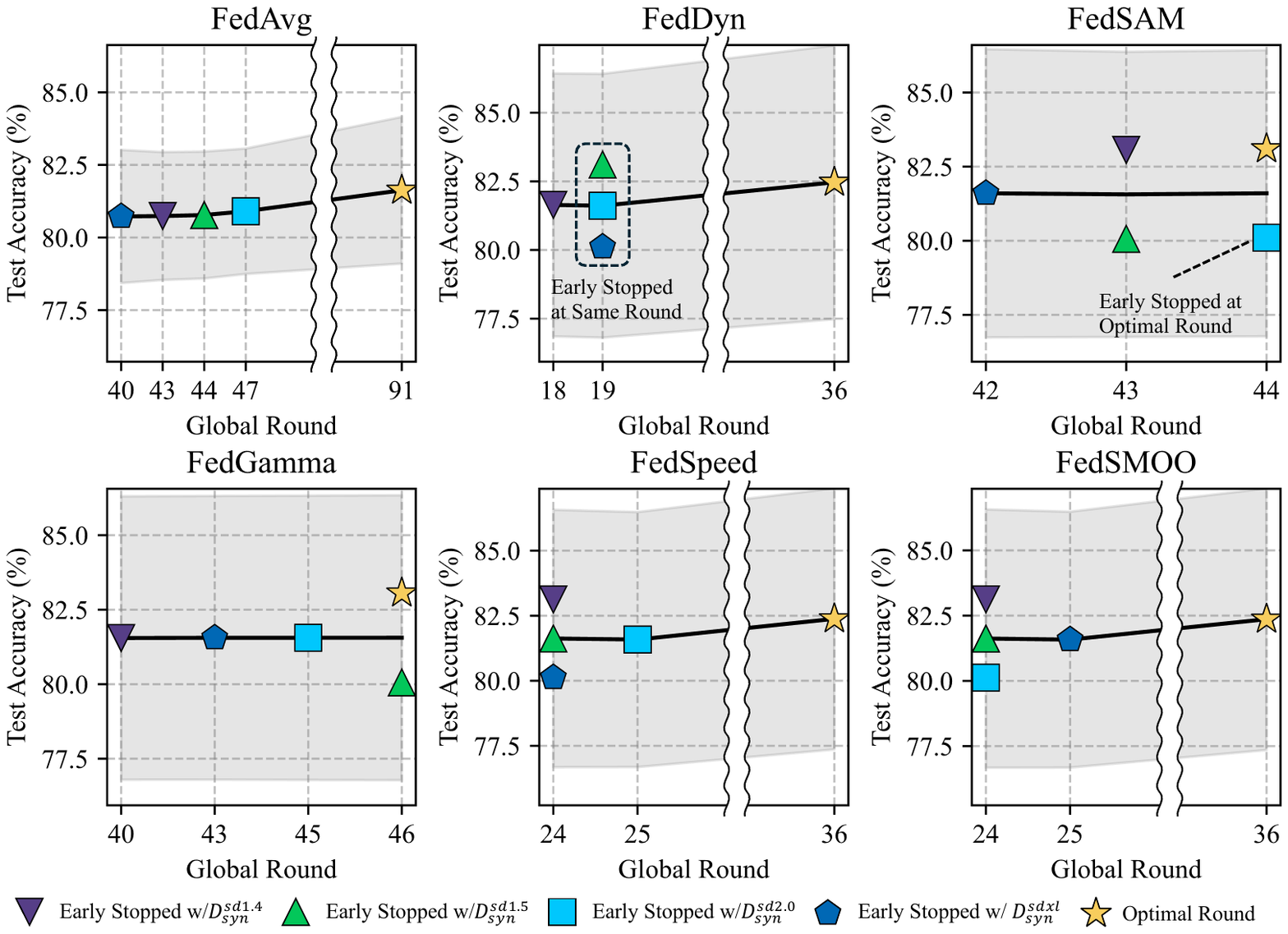}
     \caption{Comparison of early stopping performance across six FL methods using different Stable Diffusion variants. Each subplot shows the test accuracy achieved by our synthetic validation approach (colored markers: \textcolor[HTML]{563e87}{$\trianglepbfill$}, \textcolor[HTML]{03C75A}{$\trianglepafill$}, \textcolor[HTML]{00C7FD}{$\squadfill$}, \textcolor[HTML]{0068B5}{$\pentagofill$}) compared to the optimal round performance (\textcolor[HTML]{F7CD5D}{$\starletfill$}). Note that we select the best-performing configuration.}
     \label{fig:fig2}
\end{figure*}

\section{Experiment and Results}\label{sec:experiment}
\subsection{Experiment Setting}
To validate the effectiveness of our framework, we conduct experiments on the chest X-ray dataset \cite{wang2017chestx} with ResNet-18 \cite{he2016deep}, using AMD Instinct MI300X GPUs supported by AMD Developer Cloud credits.
Specifically, we partition the training dataset across $N=100$ devices using a Dirichlet distribution with $\alpha=0.1$, thereby creating non-IID nature. 
We apply our framework to recent FL methods: FedAvg \cite{mcmahan2017communication}, FedDyn \cite{acarfederated}, FedSAM \cite{qu2022generalized}, FedGamma \cite{10269141}, FedSMOO \cite{sun2023dynamic}, and FedSpeed \cite{sunfedspeed}. 
In addition, we use Stable Diffusion ($sd$) v1.4, v1.5, v2.0, and XL to generate zero-shot synthetic validation data.
In particular, we generate $D_{\text{syn}} \in \{140, 280, 700, 1400\}$ samples by creating $\eta \in \{10, 20, 50, 100\}$ images per class for $C=14$ classes, and set the early stopping patience parameter $p \in \{1, 5, 10\}$.
All experiments are run with five different random seeds, and we report the mean value. 
Additional hyperparameters and code are available at our repository\footnote{\url{https://github.com/yjlee22/earlyfl}}.

\subsection{Numerical Results}

\subsubsection{Impact of Synthetic Validation-based Early Stopping}
To investigate the efficacy of our synthetic validation-based early stopping, we compare stopping rounds and resulting test accuracy as shown in Fig.~\ref{fig:fig2}.
Using Stable Diffusion v2.0-generated validation data ($D^{sd2.0}_{syn}$), FedAvg stops at round 47 (versus optimal round 91) and achieves 80.95\% accuracy—only 0.03\% below its optimal performance.
Likewise, FedDyn under the same generator and settings converges at round 36 with 82.47\% accuracy, matching its peak performance. 
Across all six FL methods, these curves (color markers) align with their optimal stopping points (yellow markers), demonstrating that synthetic validation faithfully reflects true model behavior. 
Moreover, this configuration cuts training rounds by nearly half on average, yielding substantial savings in computation and communication. 
Thus, our framework adapts seamlessly to recent FL methods and multiple generative models.

Our broader experiments reveal that larger synthetic sample sizes and moderate patience levels yield the best results.
In particular, Stable Diffusion XL with $\eta=100$ samples per class and $p=5$ delivers the highest accuracy for FedSMOO (80.29\%) and reduces the average round gap by over 10 rounds. 
Stable Diffusion v2.0 also performs optimally with 100 samples and $p=10$, while $sd$ v1.5 at $\eta=50$ samples and $p=10$ trails by just 0.04\%. 
Conversely, using only 10 or 20 samples per class or a minimal patience of 1 leads to larger deviations from the optimal rounds and slightly lower accuracy. 
Overall, configurations with at least $\eta\geq50$ synthetic samples per class and patience of 5–10 strike the best balance between early stopping efficacy and model performance.

\begin{table}[h]
\centering
\caption{Speed‐up and accuracy deviation of synthetic validation–based early stopping under varying non‐IID degree.}
\label{tab:tab1}
\begin{tabular}{@{}llcccc@{}}
\toprule
\textbf{$\alpha$} & \textbf{Method} & \textbf{$r^*$} & \textbf{$r^*_{near}$} & \textbf{Speed-up} & \textbf{Diff (\%)} \\
\midrule
\multirow{6}{*}{0.001} & FedAvg & 70 & 57 & $\times$1.22 & -0.81 \\
 & FedDyn & 57 & 39 & $\times$1.47 & -0.96 \\
 & FedSAM & 46 & 46 & - & - \\
 & FedGamma & 100 & 58 & $\times$1.74 & -0.06 \\
 & FedSpeed & 57 & 30 & $\times$1.93 & -2.04 \\
 & FedSMOO & 57 & 30 & $\times$1.93 & -1.98 \\
\midrule
\multirow{6}{*}{0.01} & FedAvg & 57 & 47 & $\times$1.21 & -0.48 \\
 & FedDyn & 26 & 23 & $\times$1.11 & -0.01 \\
 & FedSAM & 88 & 44 & $\times$2.00 & -0.40 \\
 & FedGamma & 47 & 47 & - & - \\
 & FedSpeed & 26 & 27 & $\times$0.98 & -0.08 \\
 & FedSMOO & 26 & 26 & - & - \\
\midrule
\multirow{6}{*}{0.1} & FedAvg & 91 & 48 & $\times$1.91 & -0.72 \\
 & FedDyn & 36 & 20 & $\times$1.82 & -0.87 \\
 & FedSAM & 44 & 44 & - & - \\
 & FedGamma & 46 & 46 & - & - \\
 & FedSpeed & 36 & 25 & $\times$1.44 & -0.80 \\
 & FedSMOO & 36 & 25 & $\times$1.44 & -0.78 \\
\midrule
\multirow{6}{*}{1.0} & FedAvg & 93 & 57 & $\times$1.63 & -0.31 \\
 & FedDyn & 100 & 22 & $\times$4.55 & -0.64 \\
 & FedSAM & 81 & 50 & $\times$1.61 & -0.22 \\
 & FedGamma & 84 & 53 & $\times$1.58 & -0.02 \\
 & FedSpeed & 100 & 21 & $\times$4.67 & -0.31 \\
 & FedSMOO & 100 & 21 & $\times$4.67 & -0.35 \\
\bottomrule
\end{tabular}
\end{table}

\subsubsection{Impact of non-IID Degree}
In this experiment, we aim to demonstrate that our proposed method operates effectively regardless of data heterogeneity. 
As shown in Table~\ref{tab:tab1}, early stopping reduces training rounds by 22–74\% across non-IID degrees $\alpha\in\{0.001, 0.01, 0.1, 1.0\}$ while keeping test accuracy within almost 1\% of the optimal. 
For example, at $\alpha=0.001$, FedAvg achieves a $\times1.22$ speed‐up with only a 0.81\% drop, and at $\alpha=1.0$, FedDyn reaches a $\times4.55$ speed‐up with just a 0.64\%. 
Moreover, FedSMOO and FedSpeed exhibit consistent speed‐ups above $\times1.58$ and accuracy loss below 0.4\% across all $\alpha$, confirming robustness.
Thus, our approach functions effectively regardless of the non-IID degree, consistently maintaining near-optimal accuracy while delivering substantial computational and communication savings.
Note that, $r^*$ is the test-optimal round, serving as an upper bound for performance comparison.

Across varying synthetic generation configurations, larger sample counts and higher patience consistently yield the best performance. 
In particular, Stable Diffusion XL with $\eta=50$ samples per class and $p=10$ produces the smallest accuracy deviation (0.0–0.4\%) for FedSAM, FedSpeed, and FedSMOO.
Likewise, Stable Diffusion v2.0 with $\eta=100$ and $p=10$ is optimal for FedAvg, reducing the round gap by 12.8 rounds with only a 0.8\% accuracy loss.
In contrast, configurations with $\eta\le20$ or $p=1$ incur larger deviations of 20–75 rounds and higher accuracy loss. 
Therefore, settings with at least 50 synthetic samples per class and patience of 10 strike the ideal balance between computational savings and performance.

\begin{table}[h]
\centering
\caption{Speed‐up and accuracy deviation for synthetic validation–based early stopping using the RoentGen generator.}
\label{tab:tab2}
\begin{tabular}{@{}lcccc@{}}
\toprule
\textbf{Method} & \textbf{$r^*$} & \textbf{$r^*_{near}$} & \textbf{Speed-up} & \textbf{Diff (\%)} \\
\midrule
FedAvg & 91 & 40 & $\times$2.26 & -0.90 \\
FedDyn & 36 & 23 & $\times$1.58 & -0.85 \\
FedSAM & 44 & 40 & $\times$1.11 & -0.03 \\
FedGamma & 46 & 37 & $\times$1.24 & -0.03 \\
FedSpeed & 36 & 23 & $\times$1.54 & -0.77 \\
FedSMOO & 36 & 23 & $\times$1.59 & -0.75 \\
\bottomrule
\end{tabular}
\end{table}

\subsubsection{Ablation Study}
We perform an ablation study to evaluate the effectiveness of the domain-fine-tuned RoentGen \cite{bluethgen2024vision} in comparison with vanilla Stable Diffusion models under a non-IID setting ($\alpha=0.1$). 
As shown in Table.~\ref{tab:tab2}, the synthetic data generated with RoentGen achieves a mean speed-up factor of $\times1.55$ across FL methods, whereas the vanilla Stable Diffusion counterparts yield a factor of $\times1.43$, corresponding to an 8\% improvement in computational efficiency. 
Notably, FedAvg demonstrates a $\times2.26$ acceleration versus $\times1.91$, and FedSMOO exhibits a $1.59\times$ acceleration versus $\times1.43$. 
Despite these differences, accuracy deviations remain below 0.9\% in both scenarios, indicating negligible impact on model performance. 
These results show that domain-specific fine-tuning improves early-stopping efficacy without harming performance.
Nevertheless, ensuring its quality remains an important direction for future work.

\section{Conclusion}\label{sec:conclusion}
In this work, we propose an early stopping framework for FL that utilizes synthetic validation data generated via generative AI. 
We demonstrated that continuous evaluation on synthetic validation datasets effectively identifies near-optimal stopping rounds across state-of-the-art FL methods. 
Our experiments on non-IID chest X-ray classification tasks showed consistent speed-ups of $\times1.22\sim\times4.67$ with negligible accuracy loss. 
Furthermore, utilized fine-tuned generative AI yielded an additional 8\% improvement in computational efficiency. 
Thus, our framework enhances resource utilization and accelerates convergence in FL deployments.

\bibliographystyle{IEEEtran}
\bibliography{reference}

\end{document}